\documentclass[conference]{IEEEtran}
\IEEEoverridecommandlockouts
\usepackage{cite}
\usepackage{amsmath,amssymb,amsfonts}
\usepackage{algorithmic}
\usepackage{graphicx}
\usepackage{textcomp}
\usepackage{xcolor}
\definecolor{pythonBlue}{rgb}{0.145, 0.588, 0.945}
\usepackage{algorithm}
\usepackage{algorithmic}
\usepackage{bm}
\def\BibTeX{{\rm B\kern-.05em{\sc i\kern-.025em b}\kern-.08em
    T\kern-.1667em\lower.7ex\hbox{E}\kern-.125emX}}

\usepackage{tikz}
\usepackage{etoolbox}
\newrobustcmd*{\myVtriangle}[2]{\tikz{\filldraw[draw=#1,fill=#2] (0cm,0.2cm) --
(0.2cm,0.2cm) -- (0.1cm,0cm) -- (0cm,0.2cm);}}
\newrobustcmd*{\mythickVtriangle}[2]{\tikz{\filldraw[line width=0.3mm,draw=#1,fill=#2] (0cm,0.2cm) --
(0.2cm,0.2cm) -- (0.1cm,0cm) -- (0cm,0.2cm);}}
\newrobustcmd*{\mythickErrorVtriangle}[2]{\tikz{\filldraw[line width=0.3mm,draw=#1,fill=#2] (-0.05cm,0.05cm) --
(0.05cm,0.05cm) -- (0cm,-0.05cm) -- (-0.05cm,0.05cm);  \draw[draw=#1] (0.0cm, -0.12cm) -- (0.0cm, 0.12cm) ; \draw[draw=#1] (-0.06cm, 0.12cm) -- (0.06cm, 0.12cm); \draw[draw=#1] (-0.06cm, -0.12cm) -- (0.06cm, -0.12cm)    }}
\newrobustcmd*{\mytriangle}[2]{\tikz{\filldraw[draw=#1,fill=#2] (0.0cm,0.0cm) --
(0.2cm,0cm) -- (0.1cm,0.2cm) -- (0cm,0cm);}}
\newrobustcmd*{\mysquare}[2]{\tikz{\draw[draw=#1,fill=#2] (0cm,0cm)
rectangle (0.2cm,0.2cm)}}
\newrobustcmd*{\mythicktriangle}[2]{\tikz{\filldraw[line width=0.3mm,draw=#1,fill=#2] (0.0cm,0cm) --
(0.2cm,0cm) -- (0.1cm,0.2cm) -- (0.0cm,0cm);}}
\newrobustcmd*{\mythicksquare}[2]{\tikz{\draw[line width=0.3mm,draw=#1,fill=#2] (0cm,0cm)
rectangle (0.2cm,0.2cm)}}
\newrobustcmd*{\mybarredtriangle}[2]{\tikz{\draw[draw=#1,fill=#2] (0,0) --
(0.2cm,0) -- (0.1cm,0.2cm) -- (0cm,0cm); \draw[draw=#1] (-0.1cm, 0.07cm) -- (0.3cm, 0.07cm)}}
\newrobustcmd*{\mythickbarredtriangle}[2]{\tikz{\draw[line width=0.3mm,draw=#1,fill=#2] (0,0) --
(0.2cm,0) -- (0.1cm,0.2cm) -- (0cm,0cm); \draw[draw=#1] (-0.1cm, 0.07cm) -- (0.3cm, 0.07cm)}}
\newrobustcmd*{\mybarredsquare}[2]{\tikz{\draw[draw=#1,fill=#2] (0,0)
rectangle (0.2cm,0.2cm); \draw[draw=#1] (-0.1cm, 0.1cm) -- (0.3cm, 0.1cm)}}
\newrobustcmd*{\mythickbarredsquare}[2]{\tikz{\draw[line width=0.3mm,draw=#1,fill=#2] (0,0)
rectangle (0.2cm,0.2cm); \draw[draw=#1] (-0.1cm, 0.1cm) -- (0.3cm, 0.1cm)}}
\newrobustcmd*{\mybarredcircle}[2]{\tikz{\draw[draw=#1,fill=#2] (0,0)
circle (0.1cm); \draw[draw=#1] (-0.2cm, 0.0cm) -- (0.2cm, 0.0cm)}}
\newrobustcmd*{\mythickbarredcircle}[2]{\tikz{\draw[line width=0.3mm,draw=#1,fill=#2] (0,0)
circle (0.1cm); \draw[draw=#1] (-0.2cm, 0.0cm) -- (0.2cm, 0.0cm)}}
\newrobustcmd*{\mythickErrorcircle}[2]{\tikz{\draw[line width=0.3mm,draw=#1,fill=#2] (0,0)
circle (0.06cm); \draw[draw=#1] (0.0cm, -0.12cm) -- (0.0cm, 0.12cm) ;   \draw[draw=#1] (-0.06cm, 0.12cm) -- (0.06cm, 0.12cm); \draw[draw=#1] (-0.06cm, -0.12cm) -- (0.06cm, -0.12cm)    }}
\newrobustcmd*{\mydashedline}[1]{\tikz{\draw[draw=#1] (-0.2cm, 0.2cm) -- (-0.1cm, 0.2cm); \draw[draw=#1] (-0.0cm, 0.2cm) -- (0.1cm, 0.2cm)}}
\newrobustcmd*{\mythickcross}[1]{\tikz{\draw[line width=0.3mm,draw=#1] (0,0) --
(0.2cm,0); \draw[line width=0.3mm,draw=#1] (0.1cm,-0.1cm) -- (0.1cm,0.1cm);}}
\newrobustcmd*{\mybarredcross}[1]{\tikz{\draw[line width=0.3mm,draw=#1] (0,0) --
(0.2cm,0); \draw[line width=0.3mm,draw=#1] (0.1cm,-0.1cm) -- (0.1cm,0.1cm); \draw[draw=#1] (-0.1cm,0) -- (0.3cm,0);}}
\newrobustcmd*{\mybarredsidecross}[1]{\tikz{\draw[line width=0.3mm,draw=#1] (0,0) --
(0.3cm,0); \draw[line width=0.3mm,draw=#1] (0.1cm,-0.1cm) -- (0.2cm,0.1cm); \draw[line width=0.3mm,draw=#1] (0.2cm,-0.1cm) -- (0.1cm,0.1cm); }}
\newrobustcmd*{\myline}[1]{\tikz{\draw[draw=#1] (-0.15cm, 0.1cm) -- (0.15cm, 0.1cm);\draw[line width=0.3mm,draw=#1] (-0.0cm, 0.0cm);}}
\newrobustcmd*{\mythickline}[1]{\tikz{\draw[line width=0.3mm,draw=#1] (-0.15cm, 0.1cm) -- (0.15cm, 0.1cm);\draw[line width=0.3mm,draw=#1] (-0.0cm, 0.0cm);}}
\newrobustcmd*{\mythickdashedline}[1]{\tikz{\draw[line width=0.3mm,draw=#1] (-0.2, 0.1cm) -- (-0.1cm, 0.1cm); \draw[line width=0.3mm,draw=#1] (-0.0cm, 0.1cm) -- (0.1cm, 0.1cm); \draw[line width=0.3mm,draw=#1] (-0.0cm, 0.0cm);}}
\newrobustcmd*{\mythickdasheddottedline}[1]{\tikz{\draw[line width=0.3mm,draw=#1] (-0.22, 0.1cm) -- (-0.13cm, 0.1cm); \draw[line width=0.3mm,draw=#1] (-0.085cm, 0.1cm) -- (-0.055cm, 0.1cm); \draw[line width=0.3mm,draw=#1] (-0.01cm, 0.1cm) -- (0.08cm, 0.1cm); \draw[line width=0.3mm,draw=#1] (-0.0cm, 0.0cm);}}
\newrobustcmd*{\mycircle}[2]{\tikz{\draw[draw=#1,fill=#2] (0,0)
circle (0.1cm);}}
\newrobustcmd*{\mythickcircle}[2]{\tikz{\draw[line width=0.3mm,draw=#1,fill=#2] (0,0)
circle (0.1cm);}}
\newrobustcmd*{\mydot}[1]{\tikz{\draw[line width=0.3mm,draw=#1] (0,0)
circle (0.025cm);}}
\begin{document}

\title{Continual Adversarial Reinforcement Learning (CARL) of False Data Injection detection: forgetting and explainability\\

\thanks{$^{1}$National Renewable Energy Laboratory (NREL), Golden, CO, USA.}%
\thanks{$^{2}$Department of Electrical Engineering \& Computer Science, South Dakota State University, Brookings, SD, USA. Work completed during Pooja's internship at NREL.}%
\thanks{This work was authored in part by the National Renewable Energy Laboratory (NREL), operated by Alliance for Sustainable Energy, LLC, for the U.S. Department of Energy (DOE) under Contract No. DE-AC36-08GO28308. This work was supported by the Laboratory Directed Research and Development (LDRD) Program at NREL. The views expressed in the article do not necessarily represent the views of the DOE or the U.S. Government. The U.S. Government retains and the publisher, by accepting the article for publication, acknowledges that the U.S. Government retains a nonexclusive, paid-up, irrevocable, worldwide license to publish or reproduce the published form of this work, or allow others to do so, for U.S. Government purposes.}
\thanks{This research was performed using computational resources sponsored by the Department of Energy's Office of Energy Efficiency and Renewable Energy and located at the National Renewable Energy Laboratory.}
}

\author{\IEEEauthorblockN{Pooja Aslami$^{1,2}$, Kejun Chen$^1$, Timothy M. Hansen$^2$, Malik Hassanaly$^1$}}

\maketitle

\begin{abstract}
False data injection attacks (FDIAs) on smart inverters are a growing concern linked to increased renewable energy production. While data-based FDIA detection methods are also actively developed, we show that they remain vulnerable to impactful and stealthy adversarial examples that can be crafted using Reinforcement Learning (RL). We propose to include such adversarial examples in data-based detection training procedure via a continual adversarial RL (CARL) approach. This way, one can pinpoint the deficiencies of data-based detection, thereby offering explainability during their incremental improvement. We show that a continual learning implementation is subject to catastrophic forgetting, and additionally show that forgetting can be addressed by employing a joint training strategy on all generated FDIA scenarios. 

\end{abstract}


\section{Introduction}
\label{sec:intro}
The integration of power electronic-based distributed energy resources (DER) negatively affects the power grid inertia, increasing frequency stability issues~\cite{denholm2020inertia}. Smart inverters (SI) placed at the interface between DERs and the power grid are a promising solution to mitigate the effects of inertia loss~\cite{tamrakar2017vsm_review}. However, SIs depend on data exchanges, making their actuation policy vulnerable to False Data Injection Attacks (FDIAs), that may tamper with the exchanged data or with the control logic of the SI~\cite{nguyen2020electric}. Critically, the effects of FDIAs can be long-lived if specifically crafted to be stealthy \cite{liu2011false}. 

In view of these risks, several FDIA detection methods have been introduced and increasingly make use of data-based approaches \cite{musleh2019survey}. Such methods rely on the definition of a decision boundary that typically requires supervised training for at least some components of the FDIA detector \cite{sahu2024detection}. This exposes the detector to adversarial attacks that can exploit imperfect decision boundaries. Another approach is to proactively train a detector against adversarial examples generated as part of the training procedure \cite{chen2022self}. Here, not all adversarial examples are relevant, and important adversarial examples are the ones that defeat the detector \textit{and} induce impactful physical effects on the energy system. 

Reinforcement learning (RL) was recently shown to be a promising approach for crafting impactful adversarial examples \cite{prasad2024discovery}. One could envision a multi-agent RL (MARL) strategy where a detector agent is continuously upgraded, while an adversary agent continuously crafts new adversarial examples \cite{chen2024marl}. However, this approach is flawed with an inherent lack of explainability. Prior to deployment, the proposal of an upgraded defense strategy should be paired with a list of adversarial examples that it addresses. Within a MARL framework, a detector is trained against progressively more complex FDIA, and the types of adversarial examples exposed to the defender may not be easily accessible --- it would not be easy to know what an improved detection actually improves. Explainability strategies of RL agents have been proposed using visual explanations in the form of saliency maps \cite{heuillet2021explainability} or textual explanations, especially relevant for human-model interaction \cite{fukuchi2017autonomous}. However, no method to promote the explainability of RL policies trained against an evolving adversary has yet been proposed.

A related approach is to cast the MARL training as a \textit{continual learning} strategy. In continual learning, a model is separately trained on a sequence of tasks. In the present context, each task would be a new adversary. This approach organically pairs the defense with the adversarial examples it addresses. However, continual learning has also been shown to be vulnerable to catastrophic forgetting (CF)~\cite{goodfellow2013empirical}. Here, CF would equate to training a detector only capable against the most recent adversary, instead of training a versatile detector. 

The contributions of this work are as follows:
\begin{itemize}
    \item We demonstrate that data-based FDIA detection methods are vulnerable to impactful adversarial examples that can be constructed with an RL approach.
    \item We propose and demonstrate a continual adversarial training strategy for FDIA detection.
    \item We quantify the forgetting rate of the continual adversarial strategy, and address forgetting with the adversaries created. 
\end{itemize}


\section{Method}
\label{sec:method}

This section presents the proposed continual adversarial RL (CARL) framework. The RL environment, including the frequency dynamics model, is discussed in Sec.~\ref{sec:rlenv}. The training procedure is described in Sec.~\ref{sec:carl}.





\subsection{RL Environment}
\label{sec:rlenv}
\subsubsection{Frequency Dynamics Model}

The frequency dynamics of the power system with a primary frequency control can be modeled with the swing equation \cite{cui2022reinforcement}:
\begin{subequations}  \label{eq-swing-equation}
\begin{align}
    \dot{\theta_i} &= \omega_i,\\
    M_i\dot{\omega_i} &= p_{i}-p_{e, i}-D_i \omega_i-p^   {\text{SI}}_i,
\end{align}
\end{subequations}
\noindent where $p_{e, i} =  \sum_{j\in(\mathcal{N}\setminus\{i\})}B_{ij} \text{sin} (\theta_i - \theta_j)$, $i\in\mathcal{N}=\{1,..., n\}$ is the bus index, $\theta_i$ and $\omega_i$ are the voltage phase angle and frequency deviation of bus $i$, respectively,
$M_i$, $D_i$, $p_i$ are the inertia, damping coefficient, and net power injection of bus $i$, respectively, $p_{e,i}$ is the electric power, and $B_{i,j}$ is the susceptance of the line between bus $i$ and $j$. A linear droop controller for primary frequency control is used, i.e., $p^{\text{SI}}_i= k_i \omega_i$, where $k_i$ is the droop coefficient of the respective SI at bus $i$.

\subsubsection{False Data Injection Attack Threat Model}
\label{sec:threatmodel}
It is assumed that an adversary has gained remote access to all SIs, and can alter their control logic to induce frequency instabilities. Consistently with \cite{prasad2024discovery}, the FDIA is conducted by modifying the droop coefficient of the SI, leading to $p^{\text{SI}}_{i, t}=k^{'}_{i, t} \omega_{i, t}$ where $k'_{i,t} \in \{-1, 0, 1\}$ is the discrete altered droop coefficient for bus $i$ at time $t$. For simplification, at most one droop coefficient can be modified at a given time $t$, i.e., $\|k - k'\|_0 \leq 1$, which provides a baseline for performance analysis.
The FDIA attempts to maximize the frequency deviation, i.e., maximize $\sum_{i\in\mathcal{N}} \sum_{t\in\mathcal{T}} |\omega_i| -|\omega_i^{ref}|$, where $\omega^{ref}_i$ is the frequency deviation obtained with unaltered droop coefficients \cite{cui2022reinforcement}. Every $d$ timesteps, the adversary either selects a bus $g_t^A \in \mathcal{N}$ to attack or chooses to not attack. If it decides to attack, it can either modify the droop coefficient or temporarily mute itself. This periodic choice simplifies the definition of the detection which is described below.

\subsubsection{FDIA Detector}
The FDIA detection uses a state predictor and a classifier as described in \cite{sahu2024detection}. The state predictor is a long-short-term memory (LSTM) network that predicts the theoretical next state based on the past $d-1$ states if no FDIA occurs. Every $d$ timesteps, a neural network multi-class classifier reads the difference between the predicted and the observed states, and outputs $g_t^D \in \mathcal{N}$, the bus index thought to be under attack. While the state predictor can be trained in an unsupervised way, the classifier cannot and is vulnerable to adversarial examples. The overarching objective is to improve the classifier. The detector is successful if $g_t^D = g_t^A$, and $g_t^A$ is attacked at any point in the $d$-timestep window. The $d$-periodicity of the FDIA (Sec.~\ref{sec:threatmodel}) avoids any ambiguity about the bus under attack.

\subsection{Proposed Continual Adversarial RL (CARL) framework}
\label{sec:carl}

\begin{figure}[h]
    \centering
    \vspace{-1pt}
    \includegraphics[clip,width=0.5\textwidth]{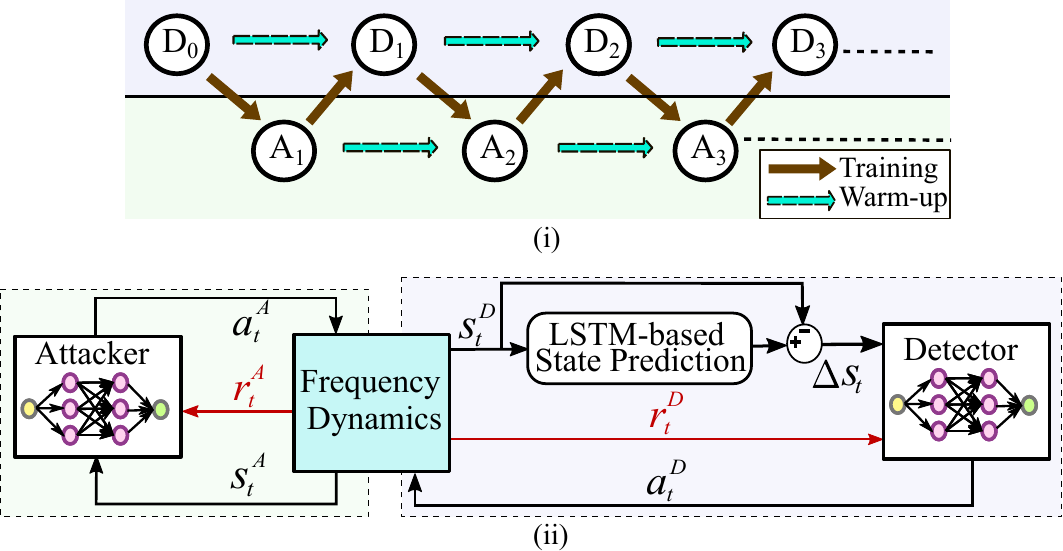}
    \centering
    \caption{\hspace{-2.7mm}(i)Schematic of the CARL process, (ii) framework of CARL process.}
    \vspace{-4pt}
    \label{fig:seqad}
\end{figure}

This section describes the proposed CARL for FDIA detection (illustrated in Fig.~\ref{fig:seqad} (i)), and includes the sequential training of an adversary RL agent and a defender RL agent. The adversary is trained to use the FDIA threat model (Sec.~\ref{sec:threatmodel}) to induce large frequency deviation while minimizing the number of successful detections from the defender agent. The defender agent is trained to maximize the number of successful detections. Below, the RL agent definitions are provided.



\noindent \subsubsection{\textbf{Adversary}} 
\noindent \paragraph{Observation} The observation $s_t^A$ of the adversary includes the frequency deviation and voltage phase angle from all buses at time $t$ and, a normalized simulation step $t'$, i.e., 
$s_t^A = [\omega_t^\mathcal{N}, \theta_t^\mathcal{N}, t']^\top$. Given that the detector is not active at every timestep, the normalized step allows the adversary to adapt its policy over time accordingly. 

\paragraph{Action} The action $a_t^A$ includes the bus under attack, tampered droop coefficient, and an attack decision variable $m_t \in\{0,1\}$, i.e., $[g_t^A, k_{i,t}', m_t]$. Here, $m_t = 1$ means that no attack  will occur at that specific timestep.

\paragraph{Reward} The reward $r_t^A$ is defined as follows:
\begin{equation}
\vspace{-10pt}
r_t^A =
\begin{cases}
    -p, & \text{if } g_t^A= g_t^D \neq -1, \forall t \in \{d, 2d, \dots\},\\
    r_t, & \text{otherwise},
\end{cases}
\label{eq:rew_ad}
\end{equation}
\noindent where, 
\begin{equation}
    r_t = c\sum_{i\in \mathcal{N}} |\omega_i|-|\omega_i^{ref}| \nonumber,
\end{equation} 

\noindent where $p>0$ is a detection penalizer and $c$ is a scaling factor that makes the penalty commensurate with the reward.


\subsubsection{\textbf{Defender}} The LSTM state predictor is not updated and only the multiclass classifier is trained.  The detection is only performed at $t \in \{d, 2d, \dots\}$. 

\paragraph{Observation} The observation $s_t^D$ for the defender includes the frequency deviation and voltage phase angle from all buses over the $d$-timestep window, i.e., $[\omega_{t-d:t}^\mathcal{N}, \theta_{t-d:t}^\mathcal{N}]^\top$.

\paragraph{Action} The action $a_t^D$ for the defender is the predicted bus index $g_t^D$.

\paragraph{Reward} The reward $r_t^D$ encourages accurate detection and is defined as:
\begin{equation}
r_t^D =
\begin{cases}
    p, & \text{if}~g_t^D= g_t^A,\\
    -p, & \text{otherwise}.
\label{eq:rew_det}
\end{cases}
\end{equation}


\subsubsection{CARL Training procedure}

The training process of $A$ and $D$ agents in the proposed CARL framework is shown in Algorithm~\ref{alg:CARL}.  Here, policy $\pi_\phi^A$ and $\pi_\theta^D$ from $A$ and $D$ are trained with separate environments $FDI^A$ and $FDI^D$, respectively. The starting point of the CARL framework is an offline FDIA detector built as described in Sec.~\ref{sec:carl}, and denoted as $D_0$ trained following Ref.~\cite{chen2024marl}. To train the classifier, throughout the episode steps, FDIAs are synthetically constructed by modifying the droop coefficient of the bus $i$ within randomly selected $\mathcal{T}_a$ fraction of steps, where $\mathcal{T}_a \in \{0.16, 0.2, 0.4, 0.6, 0.8 \}$. This synthetic attack is referred to as $A_0$ hereafter. Therefore, during an entire episode, $A_0$ targets the same bus which is not the case for the other attackers. After the first CARL iteration, and to retain memory, the agents are warm-started with the weights of the previous CARL iteration.

 \begin{algorithm}
\caption{Continual adversarial-defender training strategy}
\begin{algorithmic}[1]
\label{alg:CARL}
\FOR{$n = 1, 2, 3, \dots$ (continues as needed)}
    \STATE \textbf{\underline{Adversary Training:}} \quad
    \IF{$n = 1$}
        \STATE upload offline detector $D_0$ in $FDI^A$
    \ELSE
        \STATE upload RL detector $D_i$ in $FDI^A$ 
        \STATE $A_n \leftarrow$ warm up with policy from $A_{n-1}$
    \ENDIF
    \FOR{episode $= 1$ to $K$}
        \STATE reset $\text{FDI}^A$ and get initial state $s_0^A$
        \FOR{$t = 0$ to $\mathcal{T}-1$}
            \STATE $a_t^A \sim$ sample action from policy $\pi_{n,\phi}^A (s_t^A)$
            \STATE $s_{t+1}^A, g_t^D \leftarrow$ apply $a_t^A$ and compute state and $g_t^D$
            \STATE $r_t^A \leftarrow$ compute reward with Eq.~\ref{eq:rew_ad}, $ r^A(s_{t}^A, g_t^A, g_t^D)$
            \STATE $\phi \leftarrow $ update the parameters of $\pi_{n,\phi}^A$
            \STATE $s_t^A \leftarrow s_{t+1}^A$
        \ENDFOR
    \ENDFOR
    \STATE $\phi* \leftarrow $ save trained parameters $\phi$ in $A_n$
    \STATE \textbf{\underline{Defender Training:}} \quad
    \STATE upload RL adversary $A_n$ in $FDI^D$
    \IF{$n = 1$}
        \STATE $D_1 \leftarrow$ warm up with offline policy from $D_0$
    \ELSE
        \STATE $D_n \leftarrow$ warm up with policy from $D_{n-1}$
    \ENDIF
    \FOR{episode $= 1$ to $K$}
        \STATE reset $\text{FDI}^D$ and get initial state $s_0^D$
        \FOR{$t = 0$ to $\mathcal{T}-1$}
            \STATE $s_{t+1}^D, g_t^A \leftarrow$ compute state and $g_t^A$
            \IF{$t > 0$ \text{and} $t \mod d = 0$}
                \STATE $a_t^D \sim$ sample action from policy $\pi_{n,\theta}^D (s_t^D)$
                \STATE $r_t^A \leftarrow$ compute reward with Eq.~\ref{eq:rew_det}, $ r^D( g_t^A, g_t^D)$
            \ENDIF
            \STATE $\theta \leftarrow $ update the parameters of $\pi_{n,\theta}^D$
            \STATE $s_t^D \leftarrow s_{t+1}^D$
        \ENDFOR
    \ENDFOR
    \STATE $\theta* \leftarrow $ save trained parameters $\theta$ in $D_n$
\ENDFOR
\end{algorithmic}
\end{algorithm}

\subsection{Rehearsal Continual Adversarial RL (R-CARL)}
\label{sec:r-carl}
To address the issue of CF in continual learning \cite{goodfellow2013empirical}, several methods have been proposed such as loss regularization \cite{kirkpatrick2017overcoming,li2017learning} and model expansion strategies \cite{rusu2016progressive}. When possible, methods that use examples of old tasks when training for a new task were particularly successful \cite{lopez2017gradient,li2017learning}. Here, a rehearsal continual adversarial RL (R-CARL) approach that trains an ultimate detector $\bm{D}$ against an ensemble of adversaries is used to address the problem of CF. The same procedure can be used to train an ultimate adversary $\bm{A}$ against an ensemble of detectors.


R-CARL follows Algorithm~\ref{alg:CARL} with $n=1$ and modification to steps 4 and 21: instead of uploading a single $D$ and $A$ agent in the $FDI^D$ and $FDI^A$ environments, this approach incorporates the entire list of $D$ and $A$ agents obtained from CARL training, including $D_0$ in the defender list. The agent chosen for each environment step is sampled from the probability distribution $P$, defined in Eq.~\ref{eq:ulimateagent}.
\vspace{-4mm}



\begin{subequations}
    \label{eq:ulimateagent}
    \begin{equation}
    \bm{A} \leftarrow \text{Train against} \ \mathcal{D} \sim \mathcal{U}\{ D_0, D_1, \dots, D_N \}, 
    \end{equation}
    \begin{equation}
    \bm{D}\leftarrow \text{Train against } \mathcal{A} \sim
    \begin{cases} 
    A_0, & P(A_0) = 0.8 \\
    A_n, & P(A_i) = \frac{0.2}{N}, n \in [N],
    \end{cases}
    \end{equation}
\end{subequations}
\noindent where $N$ is the number of trained $D$ and $A$ agents from CARL training.

\section{Results}
\label{sec:results}
The system investigated is a 10-bus Kron reduced IEEE New England 39-bus~\cite{4113518}. Each episode is run for a time interval $\mathcal{T} = [0,5s]$ and is integrated with a timestep of $0.01s$ using the implementation of \cite{cui2022reinforcement}. Therefore, the output of the classifier defender contains 11 classes (one for each bus and one for a label of no attack). The RL agents were trained as proximal policy optimization stochastic agents, trained with a constant learning rate of $10^{-4}$, using the Ray library \cite{moritz2018ray}. Throughout the paper, each classifier and adversary is a neural network with 2 hidden layers and 256 neurons per layer. The detection period is $d=6$. The detection penalizer is set to $p=0.1$ the scaling factor is set to $c=0.1$.
\begin{table*}[]
\caption{Detection accuracy and frequency deviation reward for each pair of adversary and detector $(A_i, D_j)$ using CARL framework.}
\label{tab:seqad}
\centering
\begin{tabular}{|c|c|c|c|c|c|c|}
        \hline
          & $A_0$  & $A_1$ \scriptsize{\text{trained on} $D_0$}  & $A_2$ \scriptsize{\text{trained on} $D_1$}  & $A_3$ \scriptsize{\text{trained on} $D_2$} & $A_4$ \scriptsize{\text{trained on} $D_3$} &  $\bm{A}$ \scriptsize{\text{trained on} $D_{0-4}$}  \\ \hline \hline
        
        $D_0$  &  \textbf{71.6\%, 2.76}  & 3.61\%, 239.31  & 13.25\%, 233.03  & 10.84\%, 214.23 & 22.8\%, 224.5 & 15.66\%, 279.48  \\ \hline

        $D_1$ \scriptsize{\text{trained on} $A_1$}  &  23.6\%, 2.76  & \textbf{56.62\%, 239.31}  & 9.63\%, 233.03  & 7.2\%, 214.23 & 18.07\%, 224.5 & 6.02\%, 279.48  \\ \hline

        $D_2$ \scriptsize{\text{trained on} $A_2$}  & 19\%, 2.76  & 38.55\%, 239.31  & \textbf{68.6\%, 233.03} & 2.4\%, 214.23 & 13.25\%, 224.5 & 6\%, 279.48  \\ \hline 

        $D_3$ \scriptsize{\text{trained on} $A_3$}  &  16\%, 2.76  & 25.38\%, 239.31  & 43.37\%, 233.03  & \textbf{72.28\%, 214.23} & 2.4\%, 224.5 & 1.2\%, 279.48   \\ \hline 

        $D_4$ \scriptsize{\text{trained on} $A_4$}  &  13.61\%, 2.76  & 19.2\%, 239.31  & 25.3\%, 233.03  & 26.5\%, 214.23 & \textbf{83.13\%, 224.5}  & 1.2\%, 279.48 \\ \hline 

        $\bm{D}$ \scriptsize{\text{trained on} $A_{0-4}$} & 75.42\%, 2.76 & 98.79\%, 239.31 & 100\%, 233.03 & 98.799\%, 214.23 & 100\%, 224.5 & 21.68\%, 279.48 \\ \hline 

\end{tabular}
\vspace{-4pt}
\end{table*}

\subsection{Adversarial example for an offline detector}

Table~\ref{tab:seqad} shows the detection accuracy attained for each pair of $(A_i, D_j)$. Because each adversary is stochastic, the accuracy is tested on a randomly sampled realization of the adversary. For $A_0$, the detection is evaluated on a synthetic FDIA unseen at training time. In Tab.~\ref{tab:seqad}, the unscaled frequency reward ($r_t/c$ in Eq.~\ref{eq:rew_ad}) is also reported. The baseline accuracy is that of the offline defender ($D_0$) achieved against synthetic attacks ($A_0$). 

\begin{figure}[ht]
    \centering

    \hspace*{-0.7cm} 
    \includegraphics[clip,width=0.52\textwidth]{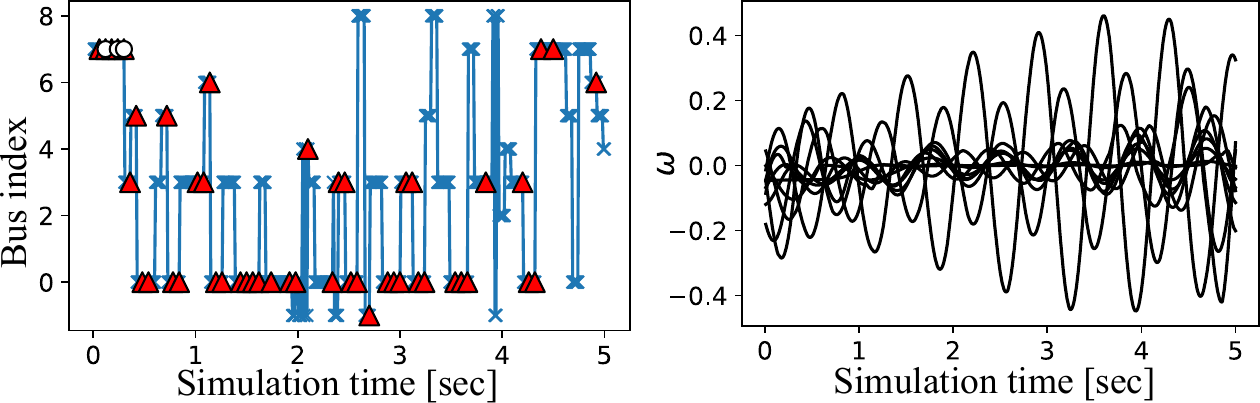}
    \caption{Left: bus under FDIA from $A_1$ (\mybarredsidecross{pythonBlue}) and corresponding successful detection from $D_0$ (\mythickcircle{black}{white}) and $D_1$ (\mythicktriangle{black}{red}). Right: time-series of frequency deviations induced for each bus induced by $A_1$.}
    \vspace{-10pt}
    \label{fig:A1}
\end{figure}

At the second CARL iteration, an adversary $A_1$ is trained to defeat $D_0$ while inducing large frequency deviations (Fig.~\ref{fig:A1} right). As shown in Tab.~\ref{tab:seqad}, $A_1$ can reduce the detection rate to $3.61\%$ while inducing large frequency deviation rewards. This shows that even detections that use a minimal amount of supervision \cite{sahu2024detection} are still vulnerable to adversarial examples that can be crafted with RL. Fig.~\ref{fig:A1} shows the time series of the attack conducted by $A_1$. $A_1$ differs from $A_0$ in that the bus targeted changes from step to step. $D_0$ only successfully detected $A_1$ at the beginning of the episode, before the first switch in the bus attacked.  

\subsection{Continual learning and forgetting}
\label{sec:forget}

Following the continual learning training sequence (Algo.~\ref{alg:CARL}), three additional detectors and adversaries are trained and the objective here is to understand how fast CF occurs. 

Tab.~\ref{tab:seqad} shows that the detection accuracy of $D_1$ on $A_0$ dropped from $71.6\%$ at the beginning of the $D_1$ training procedure to $23.6\%$. Therefore, the CARL framework is unsurprisingly subject to CF even after one CARL iteration, and warm-starting is not enough to promote knowledge retention. 

Three additional detectors ($D_2$, $D_3$, $D_4$) and three additional adversaries ($A_2$, $A_3$, $A_4$) are similarly trained, and the detection accuracies are reported in Tab.~\ref{tab:seqad}. The table diagonal (highlighted in bold), represents pairs of adversary-defenders, where the adversary action was observed by the defender. This setup leads to the highest detection accuracies. Starting from the diagonal and reading towards the left gives the backward transfer rate as defined in Ref.~\cite{lopez2017gradient}, i.e., the rate at which a defender forgets older attacks. After one iteration, the detection accuracy is reduced on average by a factor $1.7$, after two iterations by a factor $2.3$, and after three iterations by a factor $3.1$. 

Likewise, starting from the diagonal and reading downwards, the detection accuracy for each attack policy steadily decreases. After about four CARL iterations, an attack policy once addressed is almost entirely forgotten: the detection rate of $D_4$ on $A_0$ is on par with the detection rate of $D_0$ on unseen adversaries ($A_2$, $A_3$ and $A_4$).

\subsection{Addressing catastrophic forgetting with R-CARL}

Sec.~\ref{sec:forget} showed that the CARL iteration procedure was not immune to CF. To address this issue, we employ a rehearsal strategy as described in Sec.~\ref{sec:r-carl} that exposes a detector to all the available adversaries. 
The detection accuracy attained from R-CARL-trained detector $\bm{D}$ is shown in Tab.~\ref{tab:seqad}.

While not specialized in one attack, $\bm{D}$ outperforms all the other detectors, including the offline detector that was specialized to detect $A_0$. This type of accuracy improvement on a single task obtained with multi-task training \cite{rusu2015policy} suggests that adversarial examples are beneficial for improving detection rates in general. The detection accuracy on the RL adversaries is high primarily because the adversary exhibited a lack of diversity in the attack generated: therefore, a large part of the testing data was seen during training. We leave the improvement of the adversary diversity as future work. Nevertheless, the same diversity issues affected the detectors trained via continual learning ($D_1$, $D_2$, $D_3$), and the warm-started defenders could not achieve as high detection accuracy. Therefore, the rehearsal strategy outperforms the continual learning strategy, while retaining the detection capabilities on a synthetically generated set of FDIAs ($A_0$).

A similar procedure can be conducted for the adversaries by exposing one ($\bm{A}$) to a library of detectors. There again, the detection accuracies are almost always lower than all the CARL-trained adversaries (except for $D_0$ and $D_2$).

\subsection{Explainability of adversaries}

The benefit of the CARL and R-CARL approaches is that one can access the individual adversaries that are incrementally added to the knowledge pool of the detector. Here, the objective is to leverage this property to understand how adversaries change throughout the CARL iterations.

\begin{figure}[ht]
    \centering
    \vspace{-0.3cm}
    \includegraphics[clip,width=0.4\textwidth]{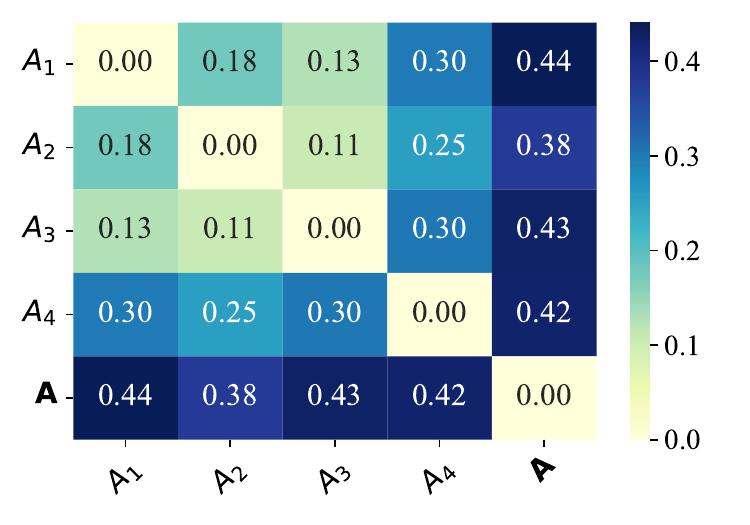}
    \centering
    \vspace{-0.44cm}
    \caption{Transition matrix differences (Eq.~\ref{eq:dist}) between pairs of adversaries.}
    \vspace{-7pt}
    \label{fig:heatmap}
\end{figure}

In a threat model (Sec.~\ref{sec:threatmodel}), every $d=6$ timesteps an adversary chooses a single bus and can either attack or not. 
The choice of the attacked bus for adversary $A_k$ is represented by a transition matrix $T_k$ where the entry of row $i$ and column $j$ is filled with the probability $P_{i\rightarrow j}$. This probability denotes the probability that if bus $i$ is targeted during one 6-timestep window, bus $j$ will be targeted during the next window. The difference $\varepsilon_{kl}$ between adversaries $A_k$ and $A_l$ can be measured as 
\begin{equation}
    \label{eq:dist}
    \varepsilon_{kl} = \frac{||T_k - T_l||}{\sqrt{||T_k|| ||T_l|| }},
\end{equation}
where $||.||$ denotes the Frobenius norm. The values of $\varepsilon_{kl}$ are shown in Fig.~\ref{fig:heatmap} for each pair of adversary and detector (shown also in the labels). Overall, while the CARL-trained adversaries ($A_{1,..4}$) behave similarly, $\bm{A}$ (trained with R-CARL) differs from them. Therefore, although the detection accuracy drops at each CARL iteration, the adversarial examples constructed minimally change from one CARL iteration to the next. This suggests that only a small change in inputs can fool data-based FDIA detectors. 

\begin{figure}[ht]
    \centering
    \vspace{-0.4cm}
    \includegraphics[clip,width=0.4\textwidth]{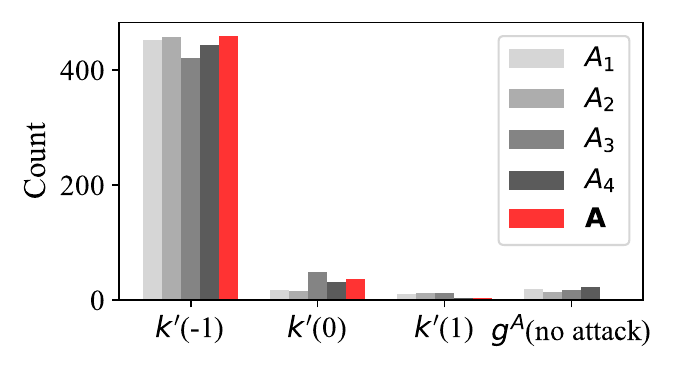}
    \centering
    \vspace{-0.45cm}
    \caption{Histogram of droop coefficient modification for all adversaries.}
    \vspace{-10pt}
    
    \label{fig:kvalue}
\end{figure}

Another way to characterize adversaries is to compute how often the droop coefficients were modified and with what value. The histogram of these decisions is shown in Fig.~\ref{fig:kvalue}. All adversaries choose to primarily employ the $k'=-1$ FDIA, which was also found to be the most impactful \cite{prasad2024discovery}. Once again, while $A_1$, $A_2$, and $A_3$ behave similarly, $A_4$ only differs in that it never chooses to use a $k'=+1$ FDIA. $\bm{A}$ differs further in that it is never muted, but instead chooses to always attack a bus. The reason why the adversary trained with R-CARL differs from the other adversaries could be the fact that the adversaries of the CARL iteration leverage the CF of the detector. By contrast, the R-CARL adversary is forced to account for the combined knowledge pool of available detectors. 

\section{Conclusions}
This work demonstrated that data-based detection of FDIA is vulnerable to impactful and stealthy adversarial attacks that can be crafted with RL. Through a continual learning approach, one can build a library of adversarial examples that can be used to create an explainable knowledge pool for the detection. Currently, we simplified the training process by focusing on a single initial condition. In the future, it would be useful to relax this assumption to induce further variabilities in the adversaries, and thereby explore the space of FDIA more extensively with few CARL iterations.


\bibliographystyle{IEEEtran}
\bibliography{references}

\end{document}